\pdfoutput=1

\documentclass[11pt]{article}

\usepackage[preprint]{acl}

\usepackage{times}
\usepackage{latexsym}
\usepackage[T1]{fontenc}
\usepackage[utf8]{inputenc}
\usepackage{microtype}
\usepackage{inconsolata}
\usepackage{graphicx}
\graphicspath{{images/}}

\usepackage{makecell}
\usepackage{tabularx}
\usepackage{ragged2e}
\usepackage{multirow}
\usepackage{booktabs}
\usepackage{arydshln}
\usepackage{amssymb}
\usepackage{pifont}
\usepackage{amsmath}
\usepackage{etoolbox}
\usepackage[most]{tcolorbox}
\usepackage{xspace}
\usepackage{xcolor}

\newtcolorbox{mybox}[1][]{
  breakable,
  arc=1mm,
  boxrule=1pt,
  colback=yellow!14,
  colframe=black!80,
  fonttitle=\bfseries,
  title=#1,
  left=1mm, right=1mm, top=1mm, bottom=1mm
}

\newcommand{\Made}{\textsc{MADE}\xspace}
\newcommand{\GaoYao}{\textsc{GaoYao}\xspace}
\newcommand{\SuperBlend}{\textsc{SuperBLEnD}\xspace}
\newcommand{\Belebele}{\textsc{Belebele}\xspace}
\newcommand{\Include}{\textsc{Include}\xspace}
\newcommand{\Mmmlu}{\textsc{MMMLU}\xspace}
\newcommand{\Mgsm}{\textsc{MGSM}\xspace}
\newcommand{\Flores}{\textsc{Flores-101}\xspace}
\newcommand{\Sage}{\textsc{SAGE}\xspace}
\newcommand{\CultureScope}{\textsc{CultureScope}\xspace}
\newcommand{\MThreeExam}{\textsc{M3Exam}\xspace}
\newcommand{\salpaca}{\textsc{S-AlpacaEval}\xspace}
\newcommand{\smtbench}{\textsc{S-MT-Bench}\xspace}

\title{MADE: Beyond Scoring via a Multilingual Agentic Diagnosing Engine for Fine-Grained Evaluation Insights}

\author{
Yilun Liu, Miao Zhang, Shimin Tao, Minggui He, Chunguang Zhao, \\
\textbf{Chenxin Liu, Li Zhang, Chen Liu, Cheng Qian, Liqun Deng,} \\
\textbf{Xiaojun Meng, Daimeng Wei} \\
Huawei, China \\
\texttt{liuyilun3@huawei.com}
}

\begin{document}
\maketitle

\begin{abstract}
Multilingual and multicultural benchmarks now cover dozens of languages and model families, but the resulting score landscapes remain metric-rich and insight-poor, necessitating fine-grained multilingual post-evaluation diagnosis. However, single LLMs and open-ended agents are easily swamped by the long, noisy diagnostic input, and no reusable taxonomy exists for it. To address this, we propose \Made{}, a \textbf{M}ultilingual \textbf{A}gentic \textbf{D}iagnosing \textbf{E}ngine that decomposes post-evaluation analysis into planning, aggregate analysis, instance-level case inspection, multilingual and cultural reflection, and grounded report synthesis. \Made{} is paired with an expert-led $54$-query $\times$ $15$-language diagnostic set, evaluated on top of a large-scale multilingual evaluation substrate ($33$ model families, $11$ benchmarks, $26$ languages, $34$ cultures, $8.66$M evaluation records). Experiments show that \Made{} outperforms the strongest shared baseline by $47\%$ in diagnosis report quality and is preferred by human multilingual experts in $87.9\%$ of pairwise comparisons. Applied with multilingual experts, \Made{} further surfaces four actionable findings on deployment, iteration, and cross-cultural pitfalls, turning benchmark score tables into model-selection and remediation guidance.

\end{abstract}

\section{Introduction}
\label{sec:introduction}
Large Language Models (LLMs) have made significant strides in serving a global user base, and the ability to operate across diverse languages and cultural contexts is now a critical measure of their inclusivity \cite{achiam2023gpt, comanici2025gemini, lai2024llms}. Reflecting this shift, the community has moved from individual translation or QA tests towards system-level multilingual benchmarks that jointly cover knowledge, reasoning, reading, generation, and cultural understanding \cite{romanouinclude, bandarkar-etal-2024-belebele, openai_mmmlu_2024, myung2024blend, rystrom2025multilingual, zhang2025culturescope, liu2026gaoyao, pomerenke2025ai}. Such benchmarks now aggregate results from dozens of languages and model families, producing rich score landscapes that report where models stand on the multilingual capability map.

\noindent\textbf{However}, these benchmarks mostly report \emph{where} models stand, while model builders still need to know \emph{why} models fail and \emph{how} to act. Three critical limitations persist:

\noindent (1)~\textbf{Existing evaluations are metric-rich but insight-poor.}\ Leaderboards expose only coarse rankings, while deployment-relevant signals stay buried under aggregate scores. As discussed in §\ref{sec:discussion}, an overall winner on \Mgsm{} may underperform a same-tier model on Swahili by $12.8$ points; a model whose overall accuracy moves by only $+0.36$pp between two releases may simultaneously fix $67$ samples and regress on $65$. Surfacing such signals today requires labour-intensive expert case analysis across languages and tasks.

\noindent (2)~\textbf{Off-the-shelf strategies struggle on long, noisy diagnostic input.}\ The diagnostic input is long and noisy---$8.66$M raw evaluation records in our experiments (about $5.2$B text tokens) spanning many benchmarks, languages, and slice axes---so a single LLM, even with a long context window, is easily swamped by irrelevant slices and cannot simultaneously plan, retrieve evidence, and self-check; conversely, open-ended general-purpose agents are free enough to skip the evidence step, drift from the analysis goal, or emit plausible but ungrounded conclusions. These weaknesses surface empirically (Table~\ref{tab:main}): single-LLM baselines reach only $4.31$ and the strongest external agent $5.45$, with evidence grounding as low as $1.97$.

\noindent (3)~\textbf{Multilingual diagnosis lacks a systematic, fine-grained taxonomy.}\ Existing error-analysis efforts~\cite{cheng2024autodetect, zhang2025beyondscore, kocmi2024preliminary} mostly target a single dataset in English; as a result, the field still lacks a reusable diagnostic-query taxonomy for multilingual post-evaluation analysis, one that covers the cross-dataset, instance-level, and iteration-level needs that multilingual model owners face.

\begin{figure*}[t]
\centering
\includegraphics[width=0.92\linewidth,trim=10 10 10 10,clip]{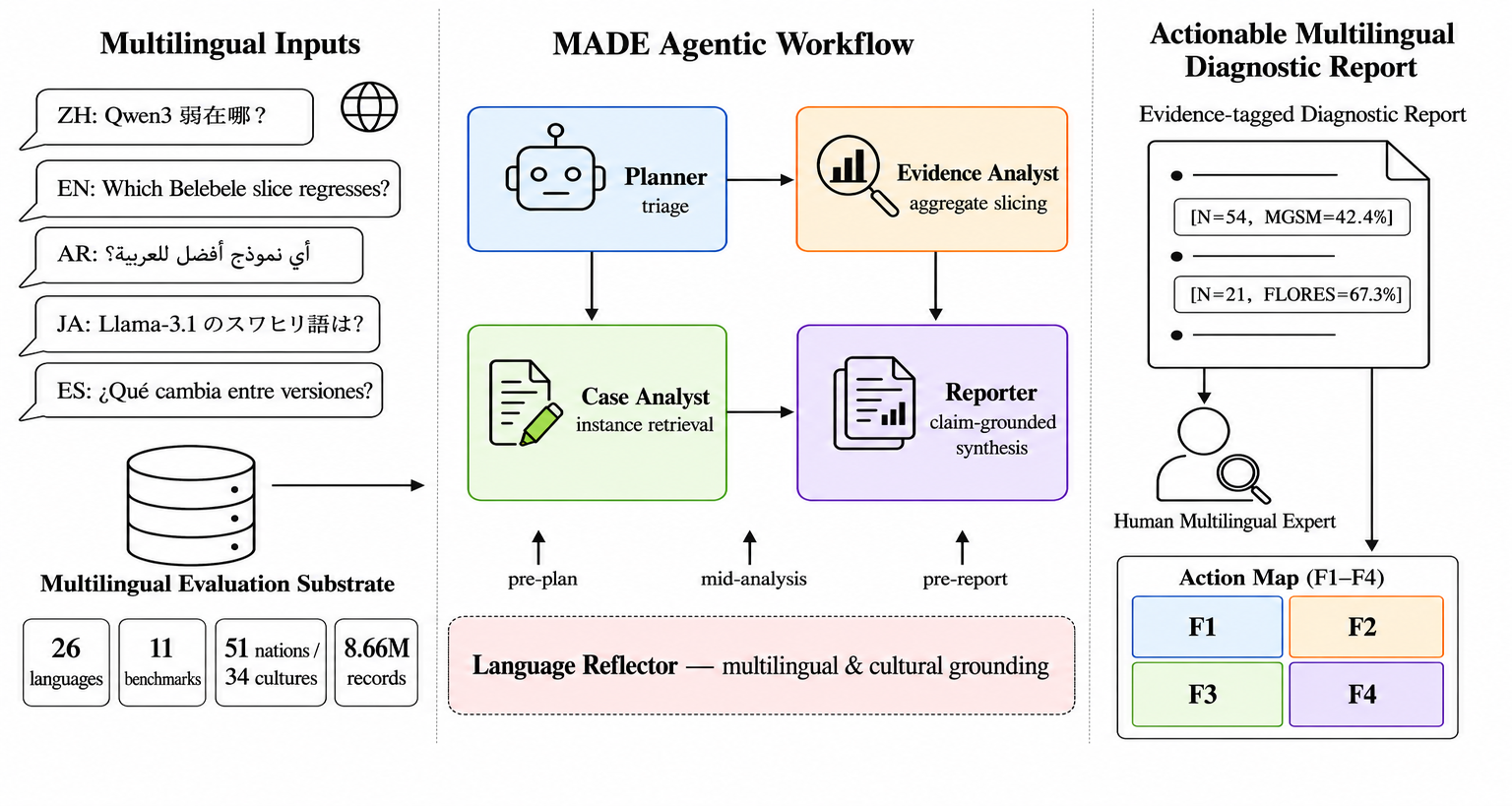}
\caption{Overview of \Made{}: multilingual user queries plus a multicultural evaluation substrate (left) are routed through five role-specialised agents (Planner, Evidence Analyst, Case Analyst, Reporter) with a Language Reflector intervening as a cross-cutting multilingual/cultural grounding layer, producing an evidence-tagged diagnostic report and an action map (right; F1--F4, §\ref{sec:discussion}).}
\label{fig:overview}
\end{figure*}

\noindent\textbf{To address these challenges}, we introduce \Made{}, a \textbf{M}ultilingual \textbf{A}gentic \textbf{D}iagnosing \textbf{E}ngine that turns large-scale benchmark outputs into fine-grained diagnostic reports (Fig.~\ref{fig:overview}), with the following innovations: \emph{(i)}~Rather than only raising report scores, \Made{}---used with human multilingual experts---turns the score landscape into an action map, yielding four actionable findings (\textit{F1--F4}, §\ref{sec:discussion}). \emph{(ii)}~To tame the long, noisy input that swamps single LLMs and the over-freedom that derails open-ended agents, \Made{} abstracts expert practice into a five-role workflow---\emph{Planner}, \emph{Evidence Analyst}, \emph{Case Analyst}, \emph{Language Reflector}, \emph{Reporter}---that both pinpoints the relevant evidence from the noise and guards against hallucination to keep every claim grounded. \emph{(iii)}~\Made{} supplies the missing taxonomy: a structured, fine-grained diagnostic query set that covers the instance-, dataset-, and iteration-level needs above. Overall, \Made{} scores $8.02$ on a 7-dimensional judge ($+2.57$ over the strongest baseline) and wins $87.9\%$ of expert pairwise comparisons. Our contributions are:
\begin{itemize}
    \setlength{\itemsep}{2pt}
    \item We \textbf{introduce} \Made{}, a multilingual agentic diagnosing engine that decomposes post-evaluation analysis into five expert agent roles, improving report quality by $47\%$ over the strongest shared baseline in a 15-language setting on real-world large-scale multilingual evaluation records (5.2B tokens from 33 model families $\times$ 26 languages $\times$ 11 benchmarks).
    \item We \textbf{demonstrate} fine-grained multilingual findings by applying \Made{} with human experts, revealing four actionable findings on deployment pitfalls, non-monotonic iteration, cultural-stance divergence, and long-tail failure modes, turning scores into action guidance.

    \item We \textbf{construct} an expert-led diagnostic query set---three evidence levels $\times$ six fine-grained categories $\times$ six query templates, instantiated as $54$ executable queries across $15$ languages---turning open-ended multilingual diagnosis needs into a reusable, balanced testbed.
\end{itemize}

\noindent In addition, we release our code and dataset to support reproducible research and a shared testbed for multilingual post-evaluation diagnosis.\footnote{\url{https://github.com/lunyiliu/MADE}.}

\section{Related Work}
\label{sec:related}
\paragraph{Multilingual and multicultural evaluation.}
Multilingual benchmarks have rapidly evolved from single-task tests \cite{goyal2022flores, shilanguage} to system-level frameworks covering knowledge, reasoning, reading, translation, subjective generation, and cultural understanding \cite{romanouinclude, bandarkar-etal-2024-belebele, openai_mmmlu_2024, myung2024blend, rystrom2025multilingual, zhang2025culturescope, liu2024omgeval, liu2026gaoyao, pomerenke2025ai}. These efforts typically produce scalar or grouped scores together with a small set of human-written aggregate findings; recent benchmarks such as~\GaoYao{} \cite{liu2026gaoyao} additionally reveal high-level patterns (\textit{e.g.}, a multilingual ``digital divide''). However, they stop short of providing a reusable engine that turns raw records into fine-grained, multilingual, and culturally grounded diagnostic reports, leaving slice-, language-, or culture-specific deployment questions to be answered by hand.

\paragraph{Post-evaluation diagnosis and error analysis.}
A complementary line begins to move beyond accuracy: error-type taxonomies for mathematical reasoning \cite{lightman2023let, shao2024deepseekmath}, MT error analysis with LLM-as-judge \cite{kocmi2024preliminary, kocmi2024findings, freitag2021experts, freitag2022results}, and pitfalls of automatic judging \cite{zouhar2024pitfalls, zheng2023judging}. More recently, automated weakness-discovery systems have explored multi-agent report generation that turns evaluation outputs into structured weakness reports \cite{cheng2024autodetect, zhang2025beyondscore}.

\paragraph{Positioning \Made{}.}
\Made{} connects the two lines above: it consumes the outputs of multilingual benchmarks and produces fine-grained diagnostic reports that name specific language, culture, and slice-level failure modes. Unlike these general-purpose weakness-discovery systems, \Made{} focuses on multilingual and multicultural, cross-dataset post-evaluation diagnosis, coupling an expert-led 15-language query set with an explicit multilingual and cultural reflection mechanism and a division of labour between aggregate-evidence and case-evidence agents that makes its conclusions more trustworthy. Our work fills this blank between large-scale multilingual evaluation and actionable, fine-grained diagnosis.

\section{Methodology}
\label{sec:method}
\subsection{Pipeline}
\label{sec:method-pipeline}

An overview of \Made{} is shown in Fig.~\ref{fig:overview}. Given a user diagnostic query and a large-scale multilingual evaluation substrate (left), \Made{} produces a structured, claim-grounded diagnostic report that human multilingual experts use as an \emph{action map} for model selection, regression auditing, and cultural-risk remediation (right). \textbf{The workflow is abstracted from how human multilingual diagnostic specialists work:} collaborating with human linguistic experts from the language service center of a top-tier corporation (on average $10{+}$ years in language services and $3{+}$ years evaluating multilingual LLMs), we observed a routine of four steps---query triage, slice-level pattern verification, representative-sample inspection, and a multilingual and cultural grounding check---which we instantiate as five role-specialised agents (expert profiles and a worked example of this routine are in Appendix~\ref{app:expert-profiles}).

\noindent\textbf{Planner.}\ The Planner outputs a structured plan specifying the diagnostic level (dataset/instance/iteration), the required analysts, the activated tool families (\textit{e.g.}, tag- or iteration-aware), and the target models, benchmarks, languages and cultures. To prevent benchmark-aware leakage, it sees only the raw query text.

\noindent\textbf{Evidence Analyst.}\ The Evidence Analyst handles dataset- and group-level evidence in a bounded ReAct loop \cite{yao2023react} over the aggregate-slicing tools of §\ref{sec:method-tools}, producing slice-level findings---rankings, gaps, and failure-rate patterns---that are bound to specific tool calls so the Reporter can later verify every quantitative claim.

\noindent\textbf{Case Analyst.}\ The Case Analyst handles instance-level diagnosis, retrieving the concrete transcripts---errors, model disagreements, degenerations, and tag-conditioned or representative cases---that aggregate statistics cannot convey. These cases let \Made{} answer instance-level queries that have no dataset-level answer (\textit{e.g.}, ``show how model $X$ fails on this slice and why''); all sample ids are drawn from a deterministic \texttt{case\_pool} for Reflector's verification to prevent example fabrication.

\noindent\textbf{Language Reflector.}\ The Language Reflector is the multilingual diagnostic specialist of \Made{}. On two orthogonal axes---cultural sensitivity and evidence grounding---it intervenes \emph{three} times: \emph{pre-plan} checks the Planner for under-specified linguistic, cultural or resource-level factors and may request a single re-plan; \emph{mid-analysis} inspects analyst findings for over-generalised claims, English-centric inferences and language/culture artefacts; \emph{pre-report} performs a final cultural and grounding audit, marking low-confidence claims and over-claims for the Reporter to downscale.

\noindent\textbf{Reporter.}\ The Reporter is the sole role that produces the final user-facing report: it receives the plan, both analysts' grounded findings and the Reflector's signals, and synthesises them into a structured, evidence-tagged diagnostic report.

\subsection{Expert Knowledge in Role Contexts}
\label{sec:method-context}

To keep the roles distinct, each role is initialised with a structured diagnostic context---written from the shared experience of the cooperating diagnostic experts rather than a free-text persona---that injects four kinds of expert knowledge: (i)~the role's objective and explicit stop conditions, (ii)~the available tool schema and the evidence expected to be bound to each tool, (iii)~multilingual and cultural caveats (resource-level reminders, common cross-lingual artefacts such as MT-induced fluency without grounding, and explicit warnings against English-centric extrapolation), and (iv)~grounding constraints that force every quantitative claim to be bound to a tool-call ledger entry. The exact per-agent injection points and previews of the core prompt components are in Appendix~\ref{app:expert-contexts}.

\subsection{Diagnostic Tools}
\label{sec:method-tools}

\Made{} exposes a small suite of \emph{deterministic} diagnostic tools across five evidence axes---language, task, model, culture, and fine-grained capability tags---so that any agent claim can be traced back to a specific tool call. Four tool families are registered: aggregate-slicing tools (group statistics, dashboards, top/bottom slice retrieval, model comparison, failure-type aggregator), instance-retrieval tools (error, disagreement, representative and degeneration cases), fine-grained capability tools that operate on \GaoYao{}-provided capability tags \cite{liu2026gaoyao}, and a cross-version delta tool for repair/regression analysis. Failure-signal tools return \emph{candidate} groups from deterministic rules (correctness transitions, response validity, degenerate repetition, cross-model disagreement, capability labels); causal labelling is left to the analysts and the Reporter, with the Language Reflector explicitly auditing such labels for over-claiming. The full registry, with representative functions, input axes, returned evidence and consuming agents per family, is in Appendix~\ref{app:tool-registry}.

\subsection{Diagnostic Taxonomy and Query Set}
\label{sec:method-taxonomy}

\Made{} is benchmarked with a diagnostic query set whose unit is a \emph{user-facing analytical query} over multilingual evaluation results. Each query asks the system to produce a diagnostic report grounded in the substrate above. To reflect real diagnostic practice as faithfully as possible and to provide a reusable resource, we first design a taxonomy that guides how the query set is generated.

\noindent\textbf{Taxonomy.}\ Queries are organised by three dimensions. The \emph{evidence level} is \emph{Dataset} (answerable from aggregate statistics), \emph{Instance} (requiring sample inspection, such as queries on fine-grained capabilities), or \emph{Iteration} (model evolution, repair and regression). The \emph{diagnostic category} is orthogonal to the evidence level: the same level of evidence can probe very different concerns, so each query also fixes \emph{which} aspect it interrogates---one of six: \emph{Task/Lang} (coverage across tasks and languages), \emph{Capability} (fine-grained sub-skills), \emph{Compliance} (instruction- and format-following), \emph{Behavior} (response patterns such as refusal or hedging), \emph{Culture} (cultural understanding and stance), and \emph{Improvement} (actionable optimisation advice). The \emph{query template} covers six practical analytical needs: single-model weakness, pairwise comparison, version/scale evolution, capability correlation, optimisation advice, and application-oriented model selection. We detail each category and the rationale for this taxonomy in Appendix~\ref{app:query-construction}.

\noindent\textbf{Construction and expansion.}\ Following the expert-led construction pattern of MIDB \cite{liu2025midb}, multilingual linguistic experts drove a five-step pipeline (taxonomy design, seed expansion, expert review, constraint filtering against the substrate, and an executability audit), yielding \textbf{54 executable diagnostic queries} authored in the experts' primary working language; these were then translated by professional language experts---with back-translation and per-(benchmark, language) audit---into a \emph{15-language diagnostic setting} spanning eight language families across East Asia, Southeast Asia, Europe and the Middle East. The full language list, per-step details and the multilingual audit protocol are in Appendix~\ref{app:query-construction}.

\noindent\textbf{Statistics.}\ The final set contains $26$ instance-level queries (representative sample inspection), $19$ dataset-level (aggregate slicing) and $9$ iteration-level (model evolution); $20/54$ ($37.0\%$) require fine-grained capability inspection and $14/54$ ($25.9\%$) are culture-heavy. Representative query types are summarised in Table~\ref{tab:query-examples}; the full joint distribution and the released query list are in Appendix~\ref{app:taxonomy-matrix}.

\begin{table}[t]
\centering\small
\setlength{\tabcolsep}{4pt}
\renewcommand{\arraystretch}{1.05}
\begin{tabular}{p{0.27\linewidth} p{0.66\linewidth}}
\toprule
Diagnostic need & Representative query \\
\midrule
Dataset-level landscape & \textit{``For model X, which languages and benchmarks form its weakest groups, and how large is the gap?''} \\
Instance-level failure & \textit{``Retrieve representative failure cases for \Belebele{} in target language $L$ and explain the error pattern.''} \\
Iteration audit & \textit{``Between two versions of the same family, which slices are fixed and which regress despite a small overall change?''} \\
Cultural \mbox{behaviour} & \textit{``When two models have similar cultural scores, do they fail through different behaviours (over-compliance, simplification, stance flattening)?''} \\
\bottomrule
\end{tabular}
\caption{Four representative diagnostic query types.}
\label{tab:query-examples}
\end{table}

\section{Experiments}
\label{sec:exp}
We evaluate \Made{} along five axes: \emph{(i)} automated evaluation of diagnostic report quality across $15$ languages (§\ref{sec:exp-main}); \emph{(ii)} human expert preference on the generated reports (§\ref{sec:exp-human}); \emph{(iii)} ablation on each component (§\ref{sec:exp-ablation}); \emph{(iv)} robustness to the underlying agent base model (§\ref{sec:exp-base}); and \emph{(v)} efficiency (Appendix~\ref{app:efficiency}). Some detailed per-dimension and per-language tables are deferred to the Appendix.

\subsection{Experimental Setup}
\label{sec:exp-setup}

\noindent\textbf{Diagnostic input.}\ The 15-language diagnostic setting introduced in §\ref{sec:method-taxonomy} contains $15 \times 54 = 810$ user queries, each grounded in a large-scale multilingual evaluation substrate spanning knowledge, reasoning, reading, translation, subjective generation, cultural understanding and exam-style tasks, across 33 model families, 11 benchmarks, 26 languages, 51 nations/areas, 34 cultures, 8.66M raw evaluation records and $\approx 5.2$B text tokens. The per-benchmark task category, language coverage and source citation are listed in Appendix~\ref{app:substrate}, and the $33$ model families in Table~\ref{tab:model-families}. All systems share the same substrate and query set.

\noindent\textbf{Evaluation criteria.}\ For automatic evaluation, each diagnostic report is scored via an LLM-as-judge on seven complementary dimensions (Requirement Fulfillment, Evidence Quality, Evidence Grounding, Readability/Structure, Multilingual Sensitivity, Diagnostic Actionability, Uncertainty Calibration; the full judge rubric with examples is in Appendix~\ref{app:judge-rubric}); the overall score is their mean. The judge receives, alongside the report, a deterministic \emph{ground-truth packet} (relevant group statistics, sample counts, sample ids and case evidence) extracted directly from the raw evaluation records by our cooperating human data analysts, and verifies whether each quantitative claim and sample reference is reproducible. We use \texttt{gemini-3.1-pro-preview} as judge, averaged over three runs per report; human evaluation uses the same factsets and dimensions (§\ref{sec:exp-human}).

\noindent\textbf{Compared systems.}\ We compare \Made{} against two classes of systems in the main multilingual experiment: (a) \emph{single-LLM} baselines---\texttt{direct}, which feeds raw evaluation-record slices directly to the LLM, and \texttt{cot}, which adds a five-role-style step-by-step CoT prompt~\cite{wei2022chain} on top; and (b) \emph{general-purpose agent frameworks}---\texttt{nanobot}~\cite{ren2026nanobot}, \texttt{opencode}~\cite{opencodeai2025opencode}, \texttt{langchain}~\cite{chase2022langchain}, \texttt{smolagents}~\cite{smolagents}, and \texttt{genericagent}~\cite{liang2026genericagent}. All baselines share the same evaluation substrate access and the same backbone, except that the agent frameworks freely decide their own analysis workflow with their own tool sets.

\noindent\textbf{Implementation Details.}\ All compared systems use a Gemini-family model (\texttt{gemini-\allowbreak{}3-\allowbreak{}flash-\allowbreak{}preview})~\cite{comanici2025gemini} as their agent backbone, with a stronger Gemini-family model (\texttt{gemini-\allowbreak{}3.1-\allowbreak{}pro-\allowbreak{}preview}) as the judge under 3-run averaging. The backbone choice is motivated by our base-swap study (§\ref{sec:exp-base}). We cap all systems' tool calls at a maximum of 15 rounds and allow up to four retries per failed (query, system) cell.

\begin{table*}[t]
\centering
\setlength{\tabcolsep}{4pt}
\resizebox{\linewidth}{!}{%
\begin{tabular}{l ccccccccccccccc cc}
\toprule
System & zh & en & ar & de & es & fr & it & ja & ko & ms & pl & pt & ru & th & tr & Avg ($\pm$std) & Valid (\%) \\
\midrule
\multicolumn{18}{l}{\textit{Single-LLM}} \\\midrule
direct & 3.82 & 4.09 & 3.90 & 3.89 & 3.82 & 4.36 & 4.32 & 4.28 & 4.38 & 4.29 & 4.71 & 4.25 & 4.47 & 4.67 & 4.26 & $4.23\pm 1.31$ & 100.0 \\
cot & 3.89 & 4.00 & 3.99 & 3.89 & 3.99 & 4.40 & 4.48 & 4.37 & 4.66 & 4.05 & 4.68 & 4.51 & 4.61 & 4.62 & 4.64 & $4.31\pm 1.25$ & 100.0 \\
\midrule
\multicolumn{18}{l}{\textit{General-purpose agent frameworks}} \\\midrule
genericagent & 4.22 & 3.26 & 2.89 & 3.25 & 3.14 & 3.70 & 3.44 & 3.79 & 3.59 & 3.55 & 3.57 & 3.44 & 3.74 & 3.45 & 3.46 & $3.50\pm 1.10$ & 100.0 \\
langchain & 4.63 & 3.41 & 3.40 & 3.67 & 3.68 & 4.05 & 4.01 & 4.18 & 3.98 & 3.77 & 4.10 & 3.94 & 4.06 & 3.73 & 4.28 & $3.93\pm 1.16$ & 99.3 \\
smolagents & 4.90 & 3.28 & 3.50 & 3.26 & 3.24 & 4.34 & 4.27 & 4.62 & 3.79 & 4.42 & 4.46 & 4.26 & 4.29 & 4.31 & 4.56 & $4.11\pm 1.23$ & 98.9 \\
nanobot & 5.59 & 4.64 & 4.68 & 4.92 & 4.87 & 5.30 & 5.42 & 5.32 & 5.27 & 4.74 & 5.57 & 5.36 & 5.49 & 4.84 & 5.23 & $5.15\pm 1.88$ & 99.6 \\
opencode & 5.86 & 4.87 & 4.89 & 5.33 & 5.09 & 5.49 & 5.43 & 5.63 & 5.71 & 5.77 & 5.56 & 5.35 & 5.54 & 5.45 & 5.63 & $5.45\pm 1.75$ & 97.0 \\
\midrule
\textbf{\Made{}} & \textbf{8.08} & \textbf{7.34} & \textbf{7.72} & \textbf{7.86} & \textbf{7.23} & \textbf{8.26} & \textbf{8.31} & \textbf{7.94} & \textbf{8.06} & \textbf{7.87} & \textbf{8.25} & \textbf{8.30} & \textbf{8.25} & \textbf{8.29} & \textbf{8.47} & $\mathbf{8.02\pm 1.04}$ & \textbf{100.0} \\
\bottomrule
\end{tabular}}
\caption{Main results on the 15-language setting (54 queries/language, 3-run judge average). \textbf{Bold} = best per column; \emph{Avg} = mean $\pm$ report-level std. Full 7-dimensional breakdown in Appendix~\ref{app:per-dim}.}
\label{tab:main}
\end{table*}

\subsection{Main Results}
\label{sec:exp-main}

Table~\ref{tab:main} reports per-language and overall automated-judge scores across the 15-language diagnostic setting. \Made{} achieves an overall judge score of $\mathbf{8.02 \pm 1.04}$, outperforming the strongest shared baseline \texttt{opencode} ($5.45 \pm 1.75$) by $+\mathbf{2.57}$ points and the strongest single-LLM baseline \texttt{cot} ($4.31 \pm 1.25$) by $+\mathbf{3.70}$ points. \Made{} wins on every single one of the 15 languages, and its report-level standard deviation ($1.04$) is also the lowest among all systems, so \Made{} is not only the most accurate but also the most stable per-report at this quality level. \Made{} is also fully robust at the cell level---$100\%$ valid in Table~\ref{tab:main}, completing every (query, language) report---whereas external agents such as \texttt{nanobot} ($0.4\%$ fail rate) and \texttt{opencode} ($3.0\%$) leave some cells unfinished even after exhausting all retries.

Per-dimension (Appendix~\ref{app:per-dim}), \Made{}'s margin is largest exactly where off-the-shelf agents are weakest---\emph{evidence grounding} ($+3.83$; $5.80$ vs.\ $1.97$ on \texttt{opencode}), \emph{uncertainty calibration} ($+3.43$) and \emph{diagnostic actionability} ($+3.35$)---while the gap on \emph{readability} is only $+0.68$: external agents write fluently but leave claims ungrounded.

\subsection{Human Evaluation}
\label{sec:exp-human}

Human expert evaluation remains necessary for multilingual diagnostic reports. Among the cooperating experts introduced in §\ref{sec:method-pipeline} (profiled in Appendix~\ref{app:expert-profiles}), we recruited senior experts with rich multilingual LLM-evaluation experience, spanning the $12$ languages for which such experts were available (\texttt{ar, de, en, fr, it, ko, pl, pt, ru, th, tr, zh}). For each of these languages we stratified-sampled $10$ of the $54$ queries at random with balanced taxonomy coverage, giving $120$ query items in total. Each item displays three blinded, randomly-shuffled reports---one from \Made{}, one from \texttt{nanobot} (the strongest stable external agent baseline with fail rate less than 1\%) and one from \texttt{cot} (the strongest single-LLM baseline)---together with the same deterministic factset shown to the automatic judge. Reviewers rated each report on the same seven dimensions used by the judge. To mitigate bias, there was no overlap between authors and reviewers, and reviewers were unaware of the source of each report.

Table~\ref{tab:human} summarises the results; per-language win rates are in Appendix~\ref{app:human-rubric}. \Made{} is not only preferred by the human experts---an $87.9\%$ human pairwise win rate, against \texttt{nanobot} $40.6\%$ and \texttt{cot} $21.5\%$---but its ranking also agrees with the automatic judge ($96.2\%$, $39.2\%$ and $14.6\%$ respectively): human and automatic preferences agree on $79.4\%$ of pairwise comparisons and pick the same best system on $80.8\%$ of items, indicating that the automatic judge is a reliable, scalable proxy for large-scale report comparison.

\begin{table}[t]
\centering\small
\setlength{\tabcolsep}{4pt}
\begin{tabular}{l rr rr}
\toprule
\multirow{2}{*}{System} & \multicolumn{2}{c}{Mean score} & \multicolumn{2}{c}{Win rate (\%)} \\
\cmidrule(lr){2-3}\cmidrule(lr){4-5}
 & Human & Auto. & Human & Auto. \\
\midrule
cot & 5.37 & 3.85 & 21.5 & 14.6 \\
nanobot & 5.93 & 5.35 & 40.6 & 39.2 \\
\midrule
\textbf{\Made{}} & \textbf{7.41} & \textbf{8.10} & \textbf{87.9} & \textbf{96.2} \\
\bottomrule
\end{tabular}
\caption{Human vs.\ automatic-judge evaluation over the 12-language human-eval set ($120$ items); human and automatic pairwise preferences agree on $79.4\%$ of comparisons and pick the same best system on $80.8\%$ of items.}
\label{tab:human}
\end{table}

\subsection{Ablation Study}
\label{sec:exp-ablation}

We ablate \Made{} on the Chinese diagnostic subset ($54$ queries), removing one component at a time. Table~\ref{tab:ablation} reports the chain. Each sub-agent contributes positively: the three largest drops come from removing the \emph{Reporter} ($-3.10$), the \emph{Evidence Analyst} ($-1.46$) and the \emph{Planner} ($-0.87$)---respectively the dedicated report synthesis, aggregate evidence and explicit routing that the workflow relies on. Removing the \emph{Case Analyst} ($-0.31$) and the \emph{Language Reflector} ($-0.19$) yield smaller drops because their contributions are targeted---instance retrieval and multilingual/cultural caveats matter on a subset of queries---yet remain positive (the Reflector specifically guards \emph{multilingual sensitivity}, which drops by $0.77$ when removed). Beyond the sub-agent chain, removing the agentic loop ($-$Agentic, i.e.\ \texttt{cot}+all tools) costs $-1.20$, and further removing tools ($-$Tool, i.e.\ \texttt{cot}) costs an additional $-3.00$. Tool grounding ($+3.00$), the agentic workflow ($+1.20$) and the full multi-agent orchestration ($+1.20$) thus contribute complementary, additive gains---none of which alone accounts for \Made{}'s margin over the baselines.

\begin{table}[t]
\centering\small
\setlength{\tabcolsep}{4pt}
\renewcommand{\arraystretch}{1.0}
\begin{tabular}{l rr}
\toprule
Variant & Overall & $\Delta$ \\
\midrule
\textbf{\Made{} (full)} & \textbf{8.08} & --- \\
\midrule
\multicolumn{3}{l}{\textit{Sub-agent ablation}} \\
\quad w/o Language Reflector & 7.89 & $-0.19$ \\
\quad w/o Case Analyst & 7.77 & $-0.31$ \\
\quad w/o Planner & 7.22 & $-0.87$ \\
\quad w/o Evidence Analyst & 6.63 & $-1.46$ \\
\quad w/o Reporter & 4.99 & $-3.10$ \\
\midrule
\multicolumn{3}{l}{\textit{Architecture ablation}} \\
\quad $-$ Agentic loop (\texttt{cot}+tools) & 6.89 & $-1.20$ \\
\quad $-$ Tools (\texttt{cot}) & 3.89 & $-4.19$ \\
\quad $-$ Tools \& CoT (\texttt{direct}) & 3.82 & $-4.27$ \\
\bottomrule
\end{tabular}
\caption{Component ablation on the Chinese diagnostic subset ($N=54$). \textbf{Bold} = full \Made{}.}
\label{tab:ablation}
\end{table}

\subsection{Base-model Sensitivity}
\label{sec:exp-base}

To check that \Made{}'s gains come from its workflow rather than from one backbone, we re-run \Made{}, \texttt{cot} and \texttt{nanobot} on the Chinese diagnostic subset under three distinct backbones from three vendors and architectures (\texttt{Gemini-3-Flash}, \texttt{Qwen3.5-122B-A10B}, \texttt{DeepSeek-V3.2}; Table~\ref{tab:base-swap}). \Made{} preserves the lead on all three: it outperforms \texttt{cot} by $+3.70$ to $+4.19$ points and \texttt{nanobot} by $+2.22$ to $+3.41$ points, with absolute scores varying in a narrow band ($8.08$--$8.49$). We adopt \texttt{Gemini-3-Flash} as the main backbone because it retains $95.2\%$ of the best observed \Made{} score ($8.08/8.49$) while running nearly $2\times$ faster per query than the strongest alternative.

\begin{table}[t]
\centering\small
\setlength{\tabcolsep}{4pt}
\begin{tabular}{l rrr r}
\toprule
Backbone & \Made{} & \texttt{cot} & \texttt{nanobot} & min/q \\
\midrule
Gemini-3-Flash & \textbf{8.08} & 3.89 & 5.59 & 4.3 \\
Qwen3.5-122B-A10B & \textbf{8.49} & 4.64 & 6.27 & 9.0 \\
DeepSeek-V3.2 & \textbf{8.33} & 4.63 & 4.92 & 9.6 \\
\bottomrule
\end{tabular}
\caption{Backbone robustness on the Chinese diagnostic subset ($N=54$); last column is the mean per-query wall-time across the three systems.}
\label{tab:base-swap}
\end{table}

\section{Diagnostic Findings}
\label{sec:discussion}
The previous section established that \Made{} produces objectively stronger diagnostic reports than strong off-the-shelf baselines. We now turn to the more important question: when multilingual experts \emph{use} \Made{} on a large-scale evaluation substrate, do they recover deployment-relevant insights that traditional leaderboard analysis cannot? We report four such findings (F1--F4).

\begin{table*}[t]
\centering\small
\setlength{\tabcolsep}{6pt}
\renewcommand{\arraystretch}{1.15}
\begin{tabular}{l r p{0.55\linewidth}}
\toprule
Iteration (lang.\ / task) & Overall $\Delta$ & Internal churn surfaced by \Made{} \\
\midrule
\makecell[tl]{\texttt{Llama-3-8B} $\to$ \texttt{Llama-3.1-8B}\\(Dutch \Include{})} & \makecell[t]{$+0.36$pp} & $67$ samples fixed, $65$ regressed ($132/551 = 23.96\%$ flip rate) \\
\midrule
\makecell[tl]{\texttt{Llama-3-8B} $\to$ \texttt{Llama-3.1-8B}\\(Portuguese \Include{})} & \makecell[t]{$-1.63$pp} & Humanities $+12.50$pp, macroeconomics $-22.73$pp, enterprise ops $-10.35$pp \\
\midrule
\makecell[tl]{\texttt{Qwen3-14B} $\to$ \texttt{Qwen3-235B-A22B}\\(Korean \Mmmlu{})} & \makecell[t]{$+13.32$pp} & Concentrated in medicine, law, history; algebra and IT stagnate or regress \\
\bottomrule
\end{tabular}
\caption{F2 iteration audits: overall delta vs.\ subdomain or sample-level churn surfaced by \Made{}.}
\label{tab:f2-iteration}
\end{table*}

\paragraph{(Meta) \Made{} turns benchmark landscapes into action maps.}\ Taken together, F1 redirects \emph{model selection} from overall rankings to slice-aware decisions; F2 redirects \emph{version iteration} from average-delta reporting to fix/regress auditing; F3 redirects \emph{cultural evaluation} from score parity to behaviour-stance auditing; and F4 redirects \emph{long-tail remediation} from a single ``low-resource is hard'' label to a remediation triage. Traditional aggregation can surface isolated symptoms in any one of these dimensions, but \Made{} routinises the path from symptom to evidence-grounded action map bound to specific languages, slices, samples, and remediation moves, suggesting that the next frontier in multilingual benchmarking is not richer leaderboards but reproducible diagnosis pipelines aligned with cultural and deployment realities.

\paragraph{(F1) Beyond Leaderboards: same-tier overall leadership does not transfer cleanly to in-context superiority.}\ A model that wins on overall accuracy may still lose on a particular language, task, or fine-grained slice that matters at deployment time. In multilingual mathematical reasoning, \texttt{Qwen3-8B} beats \texttt{Llama-3.1-8B} on \Mgsm{} overall by $+11.45$pp ($79.6$ vs.\ $68.2$); yet on Swahili the order flips, with \texttt{Llama-3.1-8B} ($55.2$) ahead of \texttt{Qwen3-8B} ($42.4$) by $+12.8$pp. The same flip recurs in reading: \texttt{Qwen3-8B} leads the 21-language \Belebele{} averages ($84.81$), yet on the German life-travel related fine-grained capability, it drops to $71.43$ while peers reach $89.80$ and $87.76$. These are not isolated anecdotes: in a same-tier compact-model audit over $301$ language-task slices (four models on \Belebele{}, \Include{}, \Mgsm{}, \Mmmlu{}), \Made{} flagged leaderboard reversals---the overall winner not being the best local choice---on $12.0\%$ of slices. We therefore treat leaderboard ranking as a starting point for model selection, not a deployment policy.

\paragraph{(F2) Beyond Monotonic Iteration: a flat overall delta between two releases can hide large internal churn.}\ Reporting an average gain conceals which subset of samples is actually fixed and which is silently regressed. Table~\ref{tab:f2-iteration} shows three representative iteration audits surfaced by \Made{}: a near-zero overall delta hiding hundreds of sample-level flips (Dutch \Include{}); an overall drop hiding a $+12.5$pp humanities gain offset by a $-22.7$pp macroeconomics loss (Portuguese \Include{}); and a large scale-up gain that concentrates in medicine, law and history while algebra and IT stagnate or regress (Korean \Mmmlu{}). A small overall multilingual delta is therefore not automatically a safe release, and \Made{}'s fix/regress map turns version iteration into actionable regression auditing.

\paragraph{(F3) Beyond Cultural Scores: equal cultural scores can hide opposite response behaviours.}\ Two models tied on a cultural slice can fail in opposite ways that a single score cannot reveal. On Chinese cross-cultural-conflict prompts from \Sage{}, \texttt{Qwen3-235B-A22B} and \texttt{Llama-3.1-405B} are statistically tied ($94.28$ vs.\ $95.29$), yet \Made{}'s case-level inspection shows they diverge in \emph{how} they respond: \texttt{Qwen3} tends to \emph{reject} the forced binary premise (e.g., countering ``Spanish cultures do not treat mountains as spiritual'' with Andean reverence)---an \emph{output failure} ($35.29\%$ of its errors: refusing the binary format)---while \texttt{Llama-3.1} \emph{complies with} and elaborates the same premise (e.g., accepting ``Chinese friendships are primarily utilitarian'')---a \emph{cognitive failure} ($78.57\%$ of its errors: stereotype compliance). The same split holds on another cultural dataset, \SuperBlend{}: \texttt{Llama-3.1} produces $50.21\%$ output failures there, whereas \texttt{GPT-4o} produces $51.01\%$ cognitive failures. Read as behavioural diagnostics rather than ideological labels, such stance differences are invisible to a single cultural score yet decisive for deployment.

\paragraph{(F4) Beyond Resource Labels: ``low-resource'' is not a single failure mode.}\ Existing multilingual benchmarks often collapse long-tail performance into a single resource label; \Made{} shows that this label hides at least three operationally different failure modes (Table~\ref{tab:f4-bucket}). A \emph{shared bottleneck}---every tier model failing the same slice (\textit{e.g.}, Yoruba \Mmmlu{} at $43$--$45\%$)---calls for corpus-side reinforcement, not a better model. A \emph{task-conditioned weakness}---the same long-tail language weak on some tasks but not others (\textit{e.g.}, \Mmmlu{} $58.39\%$ vs.\ \Belebele{} $87.33\%$ vs.\ \Mgsm{} $80.87\%$)---calls for task-specific data. A \emph{model-specific weakness}---degeneration or format instability on a single model (\textit{e.g.}, \texttt{Llama-3.1-8B} $14.05\%$ degenerate repetition vs.\ \texttt{GPT-5-chat} $0.16\%$)---calls for decoding control or model selection. The same low-resource label therefore maps to different interventions, precisely the diagnostic layer a leaderboard cannot provide.

\begin{table}[t]
\centering\small
\setlength{\tabcolsep}{2.5pt}
\renewcommand{\arraystretch}{1.05}
\begin{tabular}{p{0.20\linewidth} r p{0.34\linewidth} p{0.24\linewidth}}
\toprule
Bucket & Slices & \makecell[tl]{Representative\\evidence} & Recommended intervention \\
\midrule
Shared bottleneck & $\approx 35\%$ & Yoruba \Mmmlu{} avg $43$--$45\%$; Swahili \Mgsm{} only $42.4\%$ on \texttt{Qwen3-8B} & Corpus-side reinforcement, localised data \\
Task-conditioned weakness & $\approx 40\%$ & Same slice family: \Mmmlu{} $58.39\%$ vs.\ \Belebele{} $87.33\%$ vs.\ \Mgsm{} $80.87\%$ & Task-side data, per-task remedy \\
Model-specific weakness & $\approx 25\%$ & \texttt{Llama-3.1-8B} $14.05\%$ degenerate-repetition vs.\ \texttt{GPT-5-chat} $0.16\%$ on long-tail prompts & Decoding controls, format supervision, model selection \\
\bottomrule
\end{tabular}
\caption{F4 partitioning of long-tail failure at the diagnostic-slice level (language $\times$ task $\times$ model). Universe: $6$ languages, $4$ tasks, $7$ models, $52{,}584$ scored records.}
\label{tab:f4-bucket}
\end{table}

\section{Conclusion}
\label{sec:conclusion}
\looseness=-1
In multilingual evaluation, the bottleneck is no longer producing scores but turning them into understanding. \Made{} shows that this gap is best closed not by another leaderboard, but by treating post-evaluation \emph{diagnosis} as a first-class task---role-specialised agents that ground every claim in evidence, with multilingual and cultural reflection as an explicit step. This moves an evaluation from \emph{where} a model ranks to \emph{why} it fails and \emph{how} to act, as our four findings (F1--F4) show. As models reach ever more languages and cultures, we hope \Made{} marks a step from richer scoreboards toward reproducible, culturally accountable diagnosis---from \emph{measuring} models to \emph{improving} them. Future work includes closed-loop \emph{diagnose-then-improve} pipelines, broader benchmarks and modalities, and lighter-weight analysts to reduce reporting latency.

\clearpage
\section*{Limitations}
\label{sec:limitations}

Despite our strong empirical results and comprehensive efforts, \Made{} has several limitations:

\noindent (1) \textbf{Substrate and Benchmark Coverage.}\ \Made{} currently relies on an existing multilingual evaluation substrate---33 model families, 11 benchmarks and 26 languages---which, while large, does not cover every modality (\textit{e.g.}, vision-language), every benchmark domain (\textit{e.g.}, code or agentic tool-use), or every long-tail language. \emph{However}, \Made{} is designed as a \emph{substrate-agnostic} post-evaluation engine: its diagnostic taxonomy, tool suite and agent prompts are decoupled from the underlying benchmark adapters, so extending it to new benchmarks or modalities requires only new substrate loaders rather than a re-design of the diagnostic workflow.

\noindent (2) \textbf{Scalability of the Human-in-the-loop Pipeline.}\ The diagnostic query taxonomy, multilingual query audit, human report evaluation and finding synthesis all require multilingual expert involvement. \emph{However}, \Made{}'s goal is precisely to \emph{reduce} this human cost in steady state: experts shift from case-by-case manual diagnosis to high-level review of automatically produced reports. Human evaluation, in particular, is needed only to calibrate the automatic judge and is performed on a stratified subsample rather than on the full query set.

\noindent (3) \textbf{Computational Cost and Reporting Latency.}\ \Made{} averages $7.26$~min per report---on par with the slowest external single-agent baseline (\texttt{smolagents}, $6.24$~min) and $2$--$3\times$ slower than the most compact ones (Appendix~\ref{app:efficiency}). \emph{However}, \Made{} is intended for \emph{offline} post-evaluation diagnosis, not for online response generation, and the multi-agent decomposition stays within the runtime envelope of existing single-agent frameworks while buying a $+2.57$ to $+3.78$ point improvement in report quality. We further bound the cost via fixed role contexts, a tool-call round cap, retry budgets, batch execution and deterministic tools; future work include caching, lighter analyst models and report-template distillation to further reduce latency.

\section*{Ethics Statement}
\label{sec:ethics}

\noindent\textbf{Data and Privacy.}\ All inputs and outputs in this work originate from public or institutionally licensed multilingual benchmarks; no user-provided data, personally identifiable information or restricted-licence content is used. Diagnostic reports describe model behaviour, not human-identifiable behaviour.

\noindent\textbf{Human Evaluation.}\ The multilingual evaluators were professional language specialists who participated under standard institutional contracts; their identities are anonymised and the system identities behind each report were blinded. Annotators were free to attach optional free-text notes, which we only use for in-house qualitative analysis and do not release without consent.

\noindent\textbf{Cultural Generalisation.}\ Findings such as F3 (cultural-stance divergence) and F4 (long-tail failure modes) can be misread as essentialist claims about cultures or speaker communities. To mitigate this, \Made{}'s Reporter is contractually required to attach uncertainty calibration and caveats to every cultural claim, and the findings here are explicitly framed as model-behavioural patterns over a finite benchmark sample, not as cultural judgements about people.

\bibliography{main}

\appendix
\section{Diagnostic Query Taxonomy Matrix}
\label{app:taxonomy-matrix}

\begin{figure*}[t]
\centering
\includegraphics[width=\linewidth]{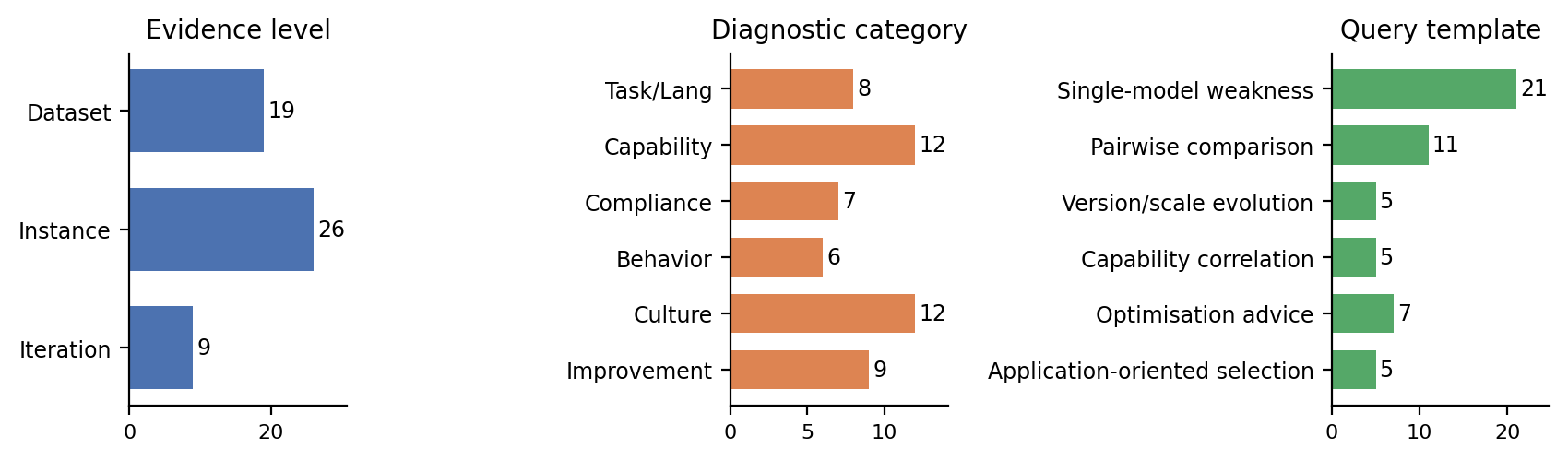}
\caption{Marginal distribution of the final $54$-query diagnostic set along the three taxonomy axes: evidence level, diagnostic category and query template.}
\label{fig:taxonomy-full}
\end{figure*}

Figure~\ref{fig:taxonomy-full} shows the marginal distribution of the final $54$-query diagnostic set along each of its three taxonomy axes---evidence level, diagnostic category and query template. The set spreads across the design space rather than concentrating in a few cells: instance-level queries form the largest evidence tier ($26$), \emph{Capability} and \emph{Culture} are the best-covered categories ($12$ each), and single-model weakness and pairwise comparison are the most frequent templates---matching the questions model owners ask most often. The main-text taxonomy of §\ref{sec:method-taxonomy} reports these marginal statistics.

\section{Baseline Configurations}
\label{app:baselines}

External agent baselines (§\ref{sec:exp-setup}) are off-the-shelf frameworks instantiated with our \texttt{gemini-3-flash-preview} backbone and the same evaluation-substrate access as \Made{}; each framework then exercises its own native tooling and analysis workflow at its own discretion: \texttt{nanobot}~\cite{ren2026nanobot}; \texttt{opencode}~\cite{opencodeai2025opencode} (SWE-CLI-style agent driven by its native \texttt{bash} loop); \texttt{langchain}~\cite{chase2022langchain} (ReAct configuration); \texttt{smolagents}~\cite{smolagents} (\texttt{CodeAgent} configuration); and \texttt{genericagent}~\cite{liang2026genericagent} (default framework configuration). We keep framework-side prompts at their published defaults to avoid prompt-engineering bias.

\section{Evaluation Substrate Coverage}
\label{app:substrate}

Table~\ref{tab:substrate} lists the 11 benchmarks underlying the diagnostic substrate (§\ref{sec:exp-setup}), with their task category and source. They span the breadth of multilingual evaluation---factual knowledge (\Mmmlu{}, \Include{}), reading comprehension (\Belebele{}), machine translation (\Flores{}), mathematical reasoning (\Mgsm{}), exam-style multi-level testing (\MThreeExam{}), subjective and multi-turn generation (\salpaca{}, \smtbench{}), and cultural understanding (\Sage{}, \SuperBlend{}, \CultureScope{}), so that the diagnosis is not tied to any single task type. Table~\ref{tab:model-families} lists the $33$ model families covered, spanning frontier proprietary and open-weight systems across $11$ developers, providing broad coverage for multilingual evaluation.

\begin{table*}[t]
\centering\small
\setlength{\tabcolsep}{6pt}
\renewcommand{\arraystretch}{1.1}
\resizebox{\linewidth}{!}{%
\begin{tabular}{l l l l}
\toprule
Benchmark & Task category & Languages / cultures covered & Source \\
\midrule
\Mmmlu{} & Knowledge & 14 languages & \cite{openai_mmmlu_2024} \\
\Include{} & Knowledge (regional) & 18 languages & \cite{romanouinclude} \\
\Belebele{} & Reading comprehension & 21 languages & \cite{bandarkar-etal-2024-belebele} \\
\Flores{} & Translation & 21 languages & \cite{goyal2022flores} \\
\Mgsm{} & Math reasoning & 11 languages & \cite{shilanguage} \\
\MThreeExam{} & Exam-style (multilevel) & 21 languages & \cite{zhang2023m3exam} \\
\salpaca{} & Subjective generation & 19 languages & \cite{liu2026gaoyao} \\
\smtbench{} & Multi-turn dialogue & 19 languages & \cite{liu2026gaoyao} \\
\Sage{} & Cross-cultural understanding & 2 languages, cross-cultural & \cite{guo2025largelanguagemodelstruly} \\
\SuperBlend{} & Cultural knowledge & English, 34 nations/cultures & \cite{liu2026gaoyao} \\
\CultureScope{} & Cultural understanding & 3 languages & \cite{zhang2025culturescope} \\
\bottomrule
\end{tabular}}
\caption{The 11 benchmarks in the evaluation substrate, by task category and source. The substrate covers 26 languages, 51 nations/areas and 34 cultures. Some benchmarks natively span more languages (\textit{e.g.}, \Flores{}-101); following GaoYao~\cite{liu2026gaoyao}, we adopt its language-slice selection---favouring languages shared across benchmarks rather than ones appearing in only a single benchmark---to keep per-language coverage balanced.}
\label{tab:substrate}
\end{table*}

\begin{table}[h]
\centering\small
\setlength{\tabcolsep}{4pt}
\renewcommand{\arraystretch}{1.2}
\begin{tabular}{p{0.20\linewidth} >{\raggedright\arraybackslash}p{0.72\linewidth}}
\toprule
Developer & Model families \\
\midrule
OpenAI & \texttt{GPT-4o}, \texttt{GPT-5-chat}, \texttt{o3}, \texttt{o4-mini} \\
Anthropic & \texttt{Claude-Sonnet-4.5} \\
Google & \texttt{Gemini-2.5-Pro}, \texttt{Gemma-3-12B-IT} \\
DeepSeek & \texttt{DeepSeek-R1}, \texttt{DeepSeek-R1-Distill-Qwen-14B}, \texttt{DeepSeek-V3.1} \\
Alibaba (Qwen) & \texttt{Qwen-7B-chat}, \texttt{Qwen-max}, \texttt{Qwen1.5-7B-Chat}, \texttt{Qwen2.5-7B-Instruct}, \texttt{Qwen3-4B}, \texttt{Qwen3-4B-Base}, \texttt{Qwen3-4B-Instruct-2507}, \texttt{Qwen3-8B}, \texttt{Qwen3-14B}, \texttt{Qwen3-32B}, \texttt{Qwen3-235B-A22B}, \texttt{Qwen3-VL-235B-A22B} \\
Meta (Llama) & \texttt{Llama-2-7b-chat}, \texttt{Meta-Llama-3-8B-Instruct}, \texttt{Llama-3.1-8B}, \texttt{Llama-3.1-70B-Instruct}, \texttt{Meta-Llama-3.1-405B-Instruct} \\
Mistral & \texttt{Ministral-8B-Instruct} \\
Zhipu & \texttt{GLM-4.6} \\
Moonshot & \texttt{Kimi-k2} \\
xAI & \texttt{Grok-3} \\
ByteDance & \texttt{Doubao-Seed-1.6} \\
Huawei & \texttt{openPangu-Ultra-MoE-718B-V1.1} \\
\bottomrule
\end{tabular}
\caption{The $33$ model families in the evaluation substrate (§\ref{sec:exp-setup}), grouped by developer.}
\label{tab:model-families}
\end{table}

\section{Expert Knowledge Contexts}
\label{app:expert-contexts}

Method §\ref{sec:method-context} summarises that each role is initialised with a structured diagnostic context injecting four kinds of expert knowledge. Table~\ref{tab:expert-contexts} lists the concrete injection points per agent: which knowledge block enters which role, what the block constrains in the role's behaviour, and where each block is sourced from.

\begin{table}[h]
\centering\small
\setlength{\tabcolsep}{3pt}
\renewcommand{\arraystretch}{1.05}
\begin{tabular}{p{0.18\linewidth} p{0.40\linewidth} p{0.34\linewidth}}
\toprule
Agent & Injected expert-knowledge blocks & Behavioural constraint enforced \\
\midrule
Planner & Diagnostic-query taxonomy primer; tool-family schema; stop conditions (no analyst forking when query is purely aggregate / case) & Picks evidence level (Dataset/Instance/Iteration); routes to analysts; bans benchmark-aware leakage from human metadata \\
Evidence Analyst & Aggregate-slicing tool schema; cross-lingual sample-size caveats; grounding contract (every claim must cite a tool call) & Stops at $\le 15$ tool calls; returns slice-level claims bound to \texttt{ToolCallLedger} entries \\
Case Analyst & Instance-retrieval tool schema; failure-mode primer (factual, cultural/pragmatic, format, degenerate, disagreement, low-sample); deterministic \texttt{case\_pool} contract & Returns representative cases with \texttt{sample\_id} guaranteed to exist; no fabricated samples \\
Language Reflector & Multilingual / multicultural caveat library; resource-tier reminders; English-centric extrapolation warnings; over-claim audit rules & At pre-plan / mid-analysis / pre-report, marks under-specified factors, over-generalisations, and low-confidence claims for downstream agents to downscale \\
Reporter & Nine-section report template; evidence-tag contract (\texttt{[N=..., metric=...]} or \texttt{[evidence=..., sample\_id=...]}); raw-data context cap ($\le$100k chars) & Synthesises the final report; every quantitative claim must be closed by an evidence tag, otherwise re-prompted \\
\bottomrule
\end{tabular}
\caption{Per-agent expert-knowledge injection in \Made{}. The contexts are not free-text personas; each block is a prompt fragment authored with multilingual specialists and pinned to a behavioural constraint that the agent's outputs are validated against.}
\label{tab:expert-contexts}
\end{table}

To show these contexts are mechanisms rather than free-text personas, we give one role-context excerpt for each agent (with implementation-specific paths and providers omitted). Each is short but exposes the role objective together with the tool, grounding, stop/retry and---where relevant---multilingual rule that constrains the agent.

\smallskip
\noindent\textbf{\emph{Planner}.}~\emph{``Triage the query into one evidence level (Dataset / Instance / Iteration) and one mode (aggregate / case / hybrid), then emit a schema-valid plan: target models, slices, tools, and which analysts to invoke. Do not fork a Case Analyst for a purely aggregate query, nor an Evidence Analyst for a purely instance query; never infer benchmark identity from human metadata. Stop once the plan validates.''}

\noindent\textbf{\emph{Evidence Analyst}.}~\emph{``Answer Dataset- and slice-level questions with aggregate-slicing tools only. Bind every claim to the tool call that produced it; if a slice's per-language sample size is below the stated threshold, flag it rather than report it as a finding. Stop after at most $15$ tool calls and return claims tied to ledger entries.''}

\noindent\textbf{\emph{Case Analyst}.}~\emph{``Retrieve representative instances (errors, model disagreements, degenerate or low-sample cases) from the deterministic case pool. Every cited case must carry a \texttt{sample\_id} that exists in the pool---never invent or paraphrase a sample. Assign each case one failure mode (factual, cultural/pragmatic, format, degenerate, disagreement) and link it to the systematic weakness it illustrates.''}

\noindent\textbf{\emph{Language Reflector}.}~\emph{``Before accepting a cross-lingual claim, check: is the gap explained by script / tokenisation rather than capability? Is the per-language sample size $\ge$ the stated threshold? Does a low-resource language inherit an English-centric assumption? Flag any claim that generalises one language's behaviour to a whole family.''}

\noindent\textbf{\emph{Reporter}.}~\emph{``Synthesise the nine-section report from the analysts' grounded claims and the Reflector's flags. Close every quantitative statement with an evidence tag; a statement without one is rejected and re-prompted. Downscale any claim the Reflector marked low-confidence, and never introduce a number absent from an analyst's ledgered evidence.''}

\section{Profiles of the Involved Human Experts and Observations on Human Diagnosing Practice}
\label{app:expert-profiles}

\Made{} is abstracted from the working routine of a multilingual diagnostic team at the language service center of a top-tier corporation. The team comes from a traditional language-service background---translation, localisation, and linguistic quality assurance, with $10+$ years of experience on average---but over the past three years the center expanded its remit from human-language services to evaluating and diagnosing multilingual large language models, giving each member $3+$ years of dedicated model-evaluation experience. This dual background is what lets them read a raw evaluation dump both statistically (which slice regresses) and pragmatically (whether a ``failure'' is a genuine capability gap or a translation/cultural artefact).

\noindent\textbf{Division of labour.}\ The team does not work as interchangeable annotators; the roles are specialised, and this specialisation is what \Made{}'s five-agent split is modelled on. A \emph{triage lead} frames the diagnostic request for each scenario---in effect posing the query---and decides what evidence and which cases are needed, and at what granularity; one or two \emph{data analysts} compute slice-level statistics and rankings; \emph{specialised reviewers} then take in those analyses, pull and read representative transcripts, and consolidate the findings into a report; and a \emph{senior multilingual reviewer} audits the others' conclusions in a multi-round, rebuttal-style manner---specifically for English-centric assumptions, low-resource sample-size issues, and over-generalised cultural claims. This multi-round human audit is the direct motivation for \Made{}'s iterative Language Reflector (§\ref{sec:method-pipeline}).

\noindent\textbf{A worked example.}\ Consider a request we observed: \emph{``Did the new release of model family $X$ actually improve Korean reading comprehension, or just its average?''} \emph{(1)~Triage.}\ The triage lead frames this as an \emph{Iteration}-level question and asks for a version-to-version comparison rather than a single ranking. \emph{(2)~Aggregation.}\ A data analyst computes per-slice deltas and finds the overall Korean score up by a few points. \emph{(3)~Case inspection.}\ Prompted by a reviewer's note that ``an average gain can hide regressions,'' a specialised reviewer retrieves the items that flipped between versions and finds that a subset of culturally-grounded reading items regressed even as factual items improved. \emph{(4)~Audit and synthesis.}\ On review, the senior reviewer flags that the regression concentrates in items requiring culture-specific inference and downscales the headline ``improved'' claim to ``improved on factual items, regressed on culturally-grounded ones.'' The deliverable is therefore an action item---audit the culturally-grounded slice before shipping---rather than a score. \Made{} inherits the \emph{spirit} of this routine rather than copying it step for step: triage, an aggregate/case split of the evidence work, a multi-round grounding audit, and a final synthesis each have a counterpart in its agents, but the correspondence is one of design intent, not literal imitation.

\section{Diagnostic Tool Registry}
\label{app:tool-registry}

Method §\ref{sec:method-tools} introduces the diagnostic tool suite. The suite follows four design principles: tools are \emph{deterministic} (same input yields the same evidence), \emph{ledgered} (every invocation is logged in a shared \texttt{ToolCallLedger} so any claim can be traced back to a specific call), \emph{claim-grounded} (failure-signal tools return candidate groups from observable rules; causal labelling is left to the agents and audited by the Language Reflector), and \emph{cross-role within \Made{}} (the Planner routes calls and the Reporter can verify any cited evidence against the ledger). Table~\ref{tab:tool-registry} lists the full paper-facing registry: tool family, a representative function name, the primary input axis the family slices over, the kind of evidence returned, which agent consumes it, and the kinds of report claims it can verifiably support.

\begin{table*}[h]
\centering\small
\setlength{\tabcolsep}{4pt}
\renewcommand{\arraystretch}{1.05}
\begin{tabular}{p{0.115\linewidth} p{0.20\linewidth} p{0.15\linewidth} p{0.165\linewidth} p{0.10\linewidth} p{0.165\linewidth}}
\toprule
Tool family & Representative function & Input axis & Returned evidence & Used by & Verifiable claim type \\
\midrule
Aggregate slicing & \texttt{group\_stats} / \texttt{dashboard} / \texttt{model\_compare} / \texttt{top\_bottom\_slices} & language $\times$ task $\times$ model $\times$ culture & per-slice mean / N / rank / cross-model delta & Evidence Analyst & ``model $X$ on slice $S$ has mean $\mu$ ($N$=...)'' \\
Instance retrieval & \texttt{error\_cases} / \texttt{disagreement\_cases} / \texttt{repr\_cases} / \texttt{degeneration\_cases} & sample $\times$ model $\times$ failure mode & sample ids drawn from deterministic \texttt{case\_pool} & Case Analyst & ``sample $i$ failed via mode $m$ on model $X$'' \\
Capability / tag analysis & \texttt{tag\_stats} / \texttt{tag\_language\_matrix} / \texttt{model\_tag\_matrix} & capability tag $\times$ language $\times$ model & per-tag accuracy / cross-tag pattern & Evidence Analyst & ``tag $t$ degrades on language $L$ for model $X$'' \\
Iteration / delta analysis & \texttt{cross\_version\_delta} / \texttt{fix\_regress\_split} & model-version pair $\times$ sample / sub-domain & fixed-sample count, regressed-sample count, sub-domain delta & Evidence Analyst / Case Analyst & ``release $X \to X'$ fixes $k_f$ samples and regresses $k_r$ samples on slice $S$'' \\
Failure-signal aggregation & \texttt{failure\_type\_stats} / \texttt{response\_validity} & sample $\times$ failure rule (correctness flip, validity, repetition, disagreement, capability label) & candidate failure groups (\emph{not} causal labels) & Evidence Analyst, audited by Reflector & ``$p\%$ of failures fall in observable bucket $b$, candidate for causal review'' \\
\bottomrule
\end{tabular}
\caption{Paper-facing diagnostic tool registry for \Made{}. All tools are deterministic, logged in a shared \texttt{ToolCallLedger}, and grounded in observable substrate fields; causal labelling is left to the agents and audited by the Language Reflector. Capability tags are treated as auxiliary descriptors and can be substituted by any comparable tagging scheme.}
\label{tab:tool-registry}
\end{table*}

\section{Diagnostic Query Set Construction and Multilingual Expansion}
\label{app:query-construction}

Method §\ref{sec:method-taxonomy} summarises the construction in one sentence; here we expand the five-step expert-led pipeline and the multilingual expansion protocol.

\noindent\textbf{Step 1 -- Taxonomy design.}\ The three-dimensional taxonomy (evidence level $\times$ diagnostic category $\times$ query template) was fixed over two rounds of author--expert discussion. The starting question was ``what does a model owner actually ask after seeing a multilingual leaderboard?''; recurring requests clustered along three orthogonal axes---\emph{how much evidence} a question needs (a dataset average, a specific failing instance, or a cross-version trajectory), \emph{which diagnostic aspect} it targets (task/language coverage, fine-grained capability, instruction compliance, response behaviour, cultural handling, or improvement advice), and \emph{what analytical shape} the answer takes (the six query templates). We kept each axis compact ($3 / 6 / 6$) so that every (level, category) pair maps to a concrete tool plan rather than an open-ended prompt; we treated the axes as saturated when two further rounds of expert requests produced no new level or category, only new instances of existing ones.

\noindent\textbf{Step 2 -- Seed expansion.}\ Grounded in the Step-1 taxonomy, the multilingual experts brainstormed candidate queries from their own diagnostic experience, each targeting a concrete analysis scenario a model owner actually faces rather than mechanically filling a cell. This ran as multiple discussion rounds until no genuinely new query was proposed---only variants of existing ones.

\noindent\textbf{Step 3 -- Expert review.}\ The multilingual specialists then reviewed each candidate on the criteria they are best placed to judge: whether any cultural premise is stated without stereotyping, whether the phrasing is free of English-centric assumptions, and whether the query carries genuine \emph{diagnostic value}---does it reflect a question practitioners actually care about in real diagnostic practice and yield actionable guidance, rather than a one-shot aggregate lookup? Candidates were retained, rewritten or merged accordingly.

\noindent\textbf{Step 4 -- Constraint filtering.}\ Candidates that violate a hard substrate constraint were then removed: a query about a (benchmark, language) pair or culture the benchmark does not cover in the substrate, or about a model the substrate does not contain, cannot be grounded in any available record and was dropped at this stage.

\noindent\textbf{Step 5 -- Executability audit.}\ The remaining queries were audited for executable answerability. We removed those whose \emph{intent} depends on capabilities the substrate cannot reliably support---long-context limit probing, strong negative-transfer causal claims (``did data $D$ cause behaviour $B$?''), and benchmark-authority judgements (``is benchmark $B$ correct?''). The audited set contains $54$ executable diagnostic queries.

\noindent\textbf{Multilingual expansion.}\ The $54$ audited queries, authored in the experts' primary working language, were translated by professional language experts into the remaining languages using a three-stage protocol: (i)~forward translation by a native speaker; (ii)~independent back-translation by a second native speaker; (iii)~per-(benchmark, language) audit to verify that the (benchmark, language, model) triples referenced in each translated query actually exist in the multilingual substrate (\textit{e.g.}, dropping a translation if it requests \Belebele{} on a language not covered by the benchmark in the substrate). This yields the 15-language diagnostic setting used throughout the experiments: \texttt{zh, en, ar, de, es, fr, it, ja, ko, ms, pl, pt, ru, th, tr}.

\section{Per-Dimension Main Results}
\label{app:per-dim}

Table~\ref{tab:per-dim-full} reports the full 7-dimensional per-system scores backing the overall numbers in main Table~\ref{tab:main}. \Made{} wins all seven dimensions; the largest gaps are on \emph{evidence grounding} ($+3.83$), \emph{uncertainty calibration} ($+3.43$), and \emph{diagnostic actionability} ($+3.35$); the smallest gap is on \emph{readability and structure} ($+0.68$).

\begin{table}[h]
\centering\small
\setlength{\tabcolsep}{3.5pt}
\resizebox{\linewidth}{!}{%
\begin{tabular}{l ccccccc c}
\toprule
System & Req. & Evi.\,Q. & Evi.\,G. & Read. & ML. & Act. & Unc. & Overall \\
\midrule
\textbf{\Made{}} & \textbf{8.35} & \textbf{7.82} & \textbf{5.80} & \textbf{8.79} & \textbf{8.48} & \textbf{8.00} & \textbf{8.87} & \textbf{8.02} \\
\texttt{opencode} & \underline{5.78} & \underline{5.46} & 1.97 & \underline{8.11} & \underline{6.75} & \underline{4.65} & \underline{5.44} & \underline{5.45} \\
\texttt{nanobot} & 5.43 & 4.89 & \underline{1.98} & 7.95 & 6.45 & 4.41 & 4.95 & 5.15 \\
\texttt{cot} & 4.03 & 3.90 & 0.82 & 7.48 & 5.99 & 4.57 & 3.42 & 4.31 \\
\texttt{direct} & 4.01 & 3.47 & 0.97 & 7.37 & 5.85 & 4.08 & 3.89 & 4.23 \\
\texttt{smolagents} & 3.84 & 3.78 & 0.37 & 7.50 & 5.66 & 3.44 & 4.17 & 4.11 \\
\texttt{langchain} & 3.72 & 3.11 & 0.55 & 7.40 & 5.62 & 3.37 & 3.72 & 3.93 \\
\texttt{genericagent} & 3.54 & 3.30 & 0.20 & 4.92 & 5.45 & 3.25 & 3.84 & 3.50 \\
\bottomrule
\end{tabular}}
\caption{Per-dimension breakdown backing main Table~\ref{tab:main}. \textbf{Bold} = best, \underline{underline} = 2nd-best unique value per column. Req.~=~Requirement Fulfillment, Evi.\,Q.~=~Evidence Quality, Evi.\,G.~=~Evidence Grounding, Read.~=~Readability/Structure, ML.~=~Multilingual Sensitivity, Act.~=~Diagnostic Actionability, Unc.~=~Uncertainty Calibration.}
\label{tab:per-dim-full}
\end{table}

This per-dimension split explains where \Made{}'s overall margin originates. The three widest gaps---\emph{evidence grounding}, \emph{uncertainty calibration} and \emph{diagnostic actionability}---are exactly the dimensions that require binding each claim to a tool call and flagging low-support statements, which off-the-shelf agents tend to skip: they retrieve little and rarely qualify their claims, so they collapse on grounding ($0.20$--$1.98$, against \Made{}'s $5.80$) even while staying fluent. The narrow \emph{readability} gap ($+0.66$) confirms the converse---every system shares a capable LLM writer, so surface fluency is not what separates them; grounded, calibrated and actionable diagnosis is.

\section{Human Evaluation Rubric and Per-Language Results}
\label{app:human-rubric}

Annotators see three blinded, randomly-shuffled reports per (query, language) item plus the deterministic factset; they rate each report along the same seven dimensions used by the automatic judge on a $[1,10]$ Likert scale and perform pairwise preferences across reports. Table~\ref{tab:human-rubric} gives the rating-dimension definitions shown to annotators. Table~\ref{tab:human-perlang} reports the per-language pairwise win rates: within each language, every (query) item yields two pairwise comparisons per system, and a system's win rate is $(W+0.5\,T)/(W+T+L)$ over all such comparisons, where $W$, $T$ and $L$ count pairwise wins, ties and losses.

\begin{table}[h]
\centering\small
\setlength{\tabcolsep}{4pt}
\renewcommand{\arraystretch}{1.05}
\resizebox{\linewidth}{!}{%
\begin{tabular}{l p{0.62\linewidth}}
\toprule
Dimension & Definition shown to the annotator \\
\midrule
Requirement Fulfillment & Does the report actually answer the user's diagnostic question? \\
Evidence Quality & Are the cited statistics / cases relevant and sufficient? \\
Evidence Grounding & Is every quantitative claim traceable to the provided factset? \\
Readability / Structure & Is the report well-organised and easy to act on? \\
Multilingual Sensitivity & Does it correctly handle language- and culture-specific nuances? \\
Diagnostic Actionability & Does it give concrete, deployable remediation guidance? \\
Uncertainty Calibration & Does it flag low-support claims instead of over-claiming? \\
\bottomrule
\end{tabular}}
\caption{Rating-dimension definitions presented to human annotators (also used by the automatic judge).}
\label{tab:human-rubric}
\end{table}

\begin{table}[h]
\centering\small
\setlength{\tabcolsep}{4pt}
\renewcommand{\arraystretch}{1.05}
\begin{tabular}{l r rrr}
\toprule
Lang & $N$ & \Made{} & nanobot & cot \\
\midrule
ar & 10 & 77.5 & 35.0 & 37.5 \\
de & 10 & 95.0 & 35.0 & 20.0 \\
en & 10 & 80.0 & 50.0 & 20.0 \\
fr & 10 & 65.0 & 55.0 & 30.0 \\
it & 10 & 90.0 & 30.0 & 30.0 \\
ko & 10 & 95.0 & 45.0 & 10.0 \\
pl & 10 & 90.0 & 45.0 & 15.0 \\
pt & 10 & 97.5 & 47.5 & 5.0 \\
ru & 10 & 90.0 & 37.5 & 22.5 \\
th & 10 & 95.0 & 15.0 & 40.0 \\
tr & 10 & 95.0 & 35.0 & 20.0 \\
zh & 10 & 85.0 & 57.5 & 7.5 \\
\bottomrule
\end{tabular}
\caption{Per-language human pairwise win rates (\%) over the $120$ evaluated items ($12$ languages, $10$ each). \Made{} has the highest win rate in every language, though its margin on \texttt{fr} ($65$ vs.\ $55$) is small.}
\label{tab:human-perlang}
\end{table}

\section{Qualitative Report Comparison}
\label{app:qualitative}

Table~\ref{tab:qualitative} shows a compact claim-level comparison of the \Made{}, \texttt{nanobot} and \texttt{cot} reports on one representative diagnostic query (Korean \Mmmlu{}, ``which sub-domains regress when scaling from Qwen3-14B to Qwen3-235B''); the factset records a $+13.32$pp overall gain that hides regressions in algebra and IT.

\begin{table}[h]
\centering\small
\setlength{\tabcolsep}{4pt}
\renewcommand{\arraystretch}{1.1}
\resizebox{\linewidth}{!}{%
\begin{tabular}{l p{0.50\linewidth} l}
\toprule
System & Representative claim in the report & Grounded? \\
\midrule
\textbf{\Made{}} & ``Overall $+13.32$pp, but algebra $-4.1$pp and IT $-2.7$pp regress \texttt{[N=120, metric=acc]}'' & Yes (factset) \\
\texttt{nanobot} & ``The larger model improves broadly, with stronger reasoning across most subjects.'' & No (no slice) \\
\texttt{cot} & ``Qwen3-235B is better than Qwen3-14B on Korean.'' & No (restates score) \\
\bottomrule
\end{tabular}}
\caption{Compact claim-level comparison on Korean \Mmmlu{} scale-up. \Made{} binds its claim to a specific sub-domain slice with an evidence tag; \texttt{nanobot} produces fluent but ungrounded prose; \texttt{cot} merely restates the headline score.}
\label{tab:qualitative}
\end{table}

The contrast is diagnostic, not stylistic. \texttt{cot} never inspects the substrate---it paraphrases the $+13.32$pp headline it was handed, so it cannot see the regression at all. \texttt{nanobot} does run analysis but stops at an aggregate impression (``improves broadly''); without binding a claim to a specific sub-domain slice, it both misses the algebra/IT regression and leaves the reader nothing to verify. \Made{} routes the same query through aggregate slicing \emph{and} case retrieval, so its Reporter states the overall gain together with the two regressing sub-domains, each closed by an evidence tag against the factset. The deployment-relevant signal---scaling up helps overall but regresses algebra and IT---survives only in the grounded pipeline; the other two reports would each lead an owner to ship the larger model without auditing the regressed slices.

\section{Automatic Judge Rubric and Prompt}
\label{app:judge-rubric}

The judge prompt presents the diagnostic query, the report under evaluation and the deterministic factset, then scores the same seven dimensions used throughout the paper. It is strict and \emph{format-agnostic}---a number is judged on whether it is correct against the factset, not on its citation style, so no system family is rewarded for its tagging conventions. The scoring core reads:
\begin{quote}\small\itshape
``You are a strict evaluator for multilingual diagnostic reports; score harshly but fairly. Verification is format-agnostic---a number is judged only on whether it is correct against the ground-truth packet, not on its citation style. Score evidence grounding $0$--$2$ when numbers appear fabricated or wrong; deduct for each factual error (accuracy, ranking, country, language, or tag) and for case ids absent from the packet. Output only a JSON object, one $0$--$10$ score per dimension, scores first.''
\end{quote}

\section{Efficiency Details}
\label{app:efficiency}

Table~\ref{tab:efficiency-full} reports per-system mean elapsed time and a quality--time index on the 15-language setting. Although \Made{} runs a five-role pipeline rather than a single agent, its $7.26$~min per query is on par with the slowest external single-agent baseline (\texttt{smolagents}, $6.24$~min) while scoring $3.91$ points higher ($8.02$ vs.\ $4.11$); against the most compact baselines it trades $2$--$3\times$ wall time for a $+2.57$ to $+3.78$ point quality gain. The multi-agent decomposition therefore buys a large quality margin while staying within the runtime envelope of existing single-agent frameworks.

\begin{table}[h]
\centering\small
\setlength{\tabcolsep}{4pt}
\begin{tabular}{l rr r}
\toprule
System & Score & Time (min/q) & QT-idx \\
\midrule
\textbf{\Made{}} & \textbf{8.02} & 7.26 & --- \\
\texttt{opencode} & 5.45 & 3.19 & 0.21 \\
\texttt{nanobot} & 5.15 & 2.76 & 0.21 \\
\texttt{smolagents} & 4.11 & 6.24 & 0.82 \\
\texttt{langchain} & 3.93 & 3.00 & 0.43 \\
\texttt{cot} & 4.31 & 2.28 & 0.27 \\
\texttt{direct} & 4.23 & 2.17 & 0.27 \\
\texttt{genericagent} & 3.50 & 1.70 & 0.20 \\
\bottomrule
\end{tabular}
\caption{Quality-time trade-off on the 15-language setting. QT-idx $= (\textsc{made\,score}-\textsc{baseline\,score})/\textsc{baseline\,score}$ divided by $\textsc{made\,time}/\textsc{baseline\,time}$. Higher values indicate more relative quality gain per unit of additional runtime; values are reported relative to each baseline. \Made{} averages $7.26$~min per report---$2.28\times$ to $3.34\times$ slower than compact baselines but $+2.57$ to $+3.78$ points higher.}
\label{tab:efficiency-full}
\end{table}

\end{document}